\documentclass[letterpaper]{article} 
\usepackage{aaai25}  
\usepackage{times}  
\usepackage{helvet}  
\usepackage{courier}  
\usepackage[hyphens]{url}  
\usepackage{graphicx} 
\usepackage{natbib}  
\usepackage{caption} 
\usepackage{algorithm}
\usepackage{algorithmic}

\usepackage{siunitx}  
\usepackage{booktabs} 
\usepackage{multirow} 

\usepackage{newfloat}
\usepackage{listings}
\DeclareCaptionStyle{ruled}{labelfont=normalfont,labelsep=colon,strut=off} 
\floatstyle{ruled}
\newfloat{listing}{tb}{lst}{}
\floatname{listing}{Listing}
\usepackage[utf8]{inputenc}

\usepackage{amsmath}
\usepackage{amsthm}
\usepackage{booktabs}
\usepackage[switch]{lineno}

\usepackage{multirow}

\usepackage{array}
\usepackage{makecell}

\usepackage{diagbox}
\usepackage{subcaption} 
\usepackage{tabularx}
\usepackage{dsfont} 
\usepackage{caption}
\usepackage{bm}

\usepackage{amssymb}

\newtheorem{theorem}{Theorem}
\newtheorem{lemma}{Lemma}
\newtheorem{definition}{Definition}
\newtheorem{prop}{Proposition}

\title{Understanding Endogenous Data Drift in Adaptive Models \\with Recourse-Seeking Users}

\author{
    Bo-Yi Liu\equalcontrib\textsuperscript{\rm 1},
    Zhi-Xuan Liu\equalcontrib\textsuperscript{\rm 1},
    Kuan Lun Chen\textsuperscript{\rm 1},
    Shih-Yu Tsai\textsuperscript{\rm 2},
    Jie Gao\textsuperscript{\rm 3},
    Hao-Tsung Yang\textsuperscript{\rm 1}
}
\affiliations {
    \textsuperscript{\rm 1}National Central University, Taiwan\\
    \textsuperscript{\rm 2}National Yang Ming Chiao Tung University, Taiwan\\
    \textsuperscript{\rm 3}Rutgers University, U.S. \\
    boyiliu1007@gmail.com,
    asd185622@gmail.com,
    milochenckl@gmail.com,\\
    shih-yu.tsai@nycu.edu.tw,
    jg1555@rutgers.edu,
    haotsungyang@gmail.com
}

\begin{document}

\maketitle

\begin{abstract}
    Deep learning models are widely used in decision-making and recommendation systems, where they typically rely on the assumption of a static data distribution between training and deployment. However, real-world deployment environments often violate this assumption. Users who receive negative outcomes may adapt their features to meet model criteria, i.e., recourse action. These adaptive behaviors create shifts in the data distribution and when models are retrained on this shifted data, a feedback loop emerges: user behavior influences the model, and the updated model in turn reshapes future user behavior. Despite its importance, this bidirectional interaction between users and models has received limited attention. In this work, we develop a general framework to model user strategic behaviors and their interactions with decision-making systems under resource constraints and competitive dynamics. Both the theoretical and empirical analyses show that user recourse behavior tends to push logistic and MLP models toward increasingly higher decision standards, resulting in higher recourse costs and less reliable recourse actions over time. To mitigate these challenges, we propose two methods—Fair-top-$k$ and Dynamic Continual Learning (DCL)—which significantly reduce recourse cost and improve model robustness. Our findings draw connections to economic theories, highlighting how algorithmic decision-making can unintentionally reinforce a higher standard and generate endogenous barriers to entry.
\end{abstract}

\begin{links}
    \link{Code}{https://github.com/asai-lab/Understanding-Endogenous-Data-Drift-in-Adaptive-Models-with-Recourse-Seeking-Users.git}
\end{links}

\section{Introduction}
Deep learning has emerged as a fundamental tool in decision-making and recommendation systems, typically structured into two main phases: \emph{training} and \emph{prediction}~\cite{zhang2019deep}. In a standard binary classification scenario, during the training phase, the model captures patterns from historical data. In the prediction phase, the trained model is deployed within a real-time system—such as YouTube’s recommendation engine or e-commerce platforms—to drive decisions or recommendations~\cite{qin2020binary,kirdemir2021examining,zhou2020product,shankar2017deep}.

This paradigm relies on the assumption that the data distribution remains unchanged post-deployment. However, in practice, deployed systems often influence the data they consume. For example, users receiving unfavorable outcomes (e.g., rejection) may engage in \emph{recourse actions}: modifying their attributes to better align with the model's criteria, typically at minimal cost~\cite{karimi2022survey,o2021multi,nguyen2023feasible,poyiadzi2020face,yadav2021low,venkatasubramanian2020philosophical}. A prominent example of this phenomenon is the proliferation of websites and articles that offer strategies to ``beat'' an algorithm, teaching users how to exploit their mechanics on recommendation systems such as Google, YouTube, Facebook, and so on~\cite{macdonald2023actually,klug2021trick}. 

However, recourse actions tend to be strategically narrow rather than diverse, for two main reasons. First, available ``slots''—such as job positions, loan approvals, or top-ranked recommendation slots—are inherently limited and highly competitive, forcing users to tailor their efforts toward aligning with the system’s criteria rather than exploring varied improvement paths.~\cite{herlocker2004evaluating,hennig2012can}. Second, users tend to focus on modifying low-cost, high-impact features. For example, optimizing for engagement metrics like clicks and likes because such changes offer relatively high returns with minimal effort, rather than investing in more substantive improvements~\cite{chen2023learning,estornell2023incentivizing}. Although these behaviors may not be explicitly dishonest or rule-breaking, they can nevertheless generate unintended and potentially harmful distortions in the data. Indeed, studies have shown that social media recommendation systems can amplify polarization and emotionally charged content, further undermining the assumption of a static data distribution post-deployment~\cite{chitra2020analyzing,chen2023learning}.

While some research addresses the influence of user recourse behavior and data drift in deployed models~\cite{hardt2025performativepredictionpastfuture,altmeyer2023endogenous}, few works explore how such recourse influences model evolution when the system adapts to new data. Our assumption is that it would amplify the deviated direction and becomes a vicious circle, creating a feedback loop: behaviors adapt to the model, the model retrains on these behaviors, thus reinforcing the skew. As an example in video recommendation, Yuval Noah Harari~\cite{harari2024nexus} explains that videos with extremist content tend to drive higher user engagement. Thus, a recommendation system designed to maximize user retention may encourage and promote such content in the end. Empirical evidence supports this claim. A systematic review shows that 21 out of 23 recent studies implicated YouTube's recommendation system in promoting problematic content pathways~\cite{yesilada2022systematic}. These findings underscore the bidirectional influence between user behavior and model updates. 

\textbf{In this work}, we investigate the interaction between system deployment and user recourse behavior, and how this dynamic affects both model updates and user features over time—referred to as model shift and data shift~\cite{hardt2016strategic}. This interaction is inherently bidirectional: the deployed model influences users’ recourse actions, leading to changes in the data distribution (data drift), and in turn, this data drift causes the model to update (model shift), completing a feedback loop. 
To the best of our knowledge, this is the first work to discuss the interaction in the long run. Two prior studies are most relevant to our setting. One focuses on the resource competition between recourse users when calculating the recourse actions~\cite{fonseca2023setting}. The other explores recourse algorithms and conditions that make the model shift but does not investigate the bidirectional long-term effects and properties triggered by the interactions between the system and users~\cite{altmeyer2023endogenous}.

To address this gap, we develop a framework that explicitly models the full interaction loop: \textit{model deployment} $\rightarrow$ \textit{user response (i.e., recourse actions)} $\rightarrow$ \textit{model update}. A key observation in this setting is that the system operates under limited resources and cannot accommodate all users, even when many improve their features. As a result, the system must update its decision criteria based on the new, competitive data distribution. Our study focuses on two central questions:

\begin{enumerate}
\item \textit{Under limited resources, how do recourse users influence the system and drive long-term model updates?}
\item \textit{When model updates are driven by recourse users, how do these changes affect future users and their recourse behavior over time?}
\end{enumerate}

\textbf{Our Contributions:}
We examine the iterative dynamics between user recourse behavior and model updates. Our key findings are summarized as follows:

\begin{enumerate}
\item We provide a theoretical analysis showing that logistic models tend to evolve toward a higher decision standard over time. This trend is further supported by simulation results across a range of settings.
\item We show that a higher decision standard leads to increased recourse costs and results in less reliable, more failure-prone recourse actions.
\item To address these challenges, we propose two strategies—Fair-top-$k$ and Dynamic Continual Learning (DCL)—which effectively mitigate the identified issues and enhance model robustness.
\end{enumerate}

Notably, the first two findings can be related to multiple economic principles. For instance, competitive environments often push systems toward producing higher-quality or higher-standard outcomes—analogous to the model adopting a stricter decision boundary in our setting~\cite{bikker2002competition, ezrachi2015curious}. Furthermore, such dynamics can generate endogenous barriers to entry, making it more difficult for new or less advantaged users to succeed~\cite{mcafee2004barrier, pehrsson2009barriers}. This directly relates to our second finding: as the model standard rises, recourse actions become increasingly costly and are more likely to fail over time.

Lastly, our proposed methods show strong empirical results. Fair-top-$k$ and DCL significantly alleviate the higher standard issue, improving performance by approximately 50\% across all rounds. Recourse costs are reduced by around 70\%, and balanced accuracy remains stable, reaching approximately 90\% in the long run—demonstrating improved robustness compared with other methods.

\begin{figure*}[ht]
    \centering
    \begin{subfigure}[t]{0.32\textwidth}
        \centering
        \includegraphics[width=\linewidth]{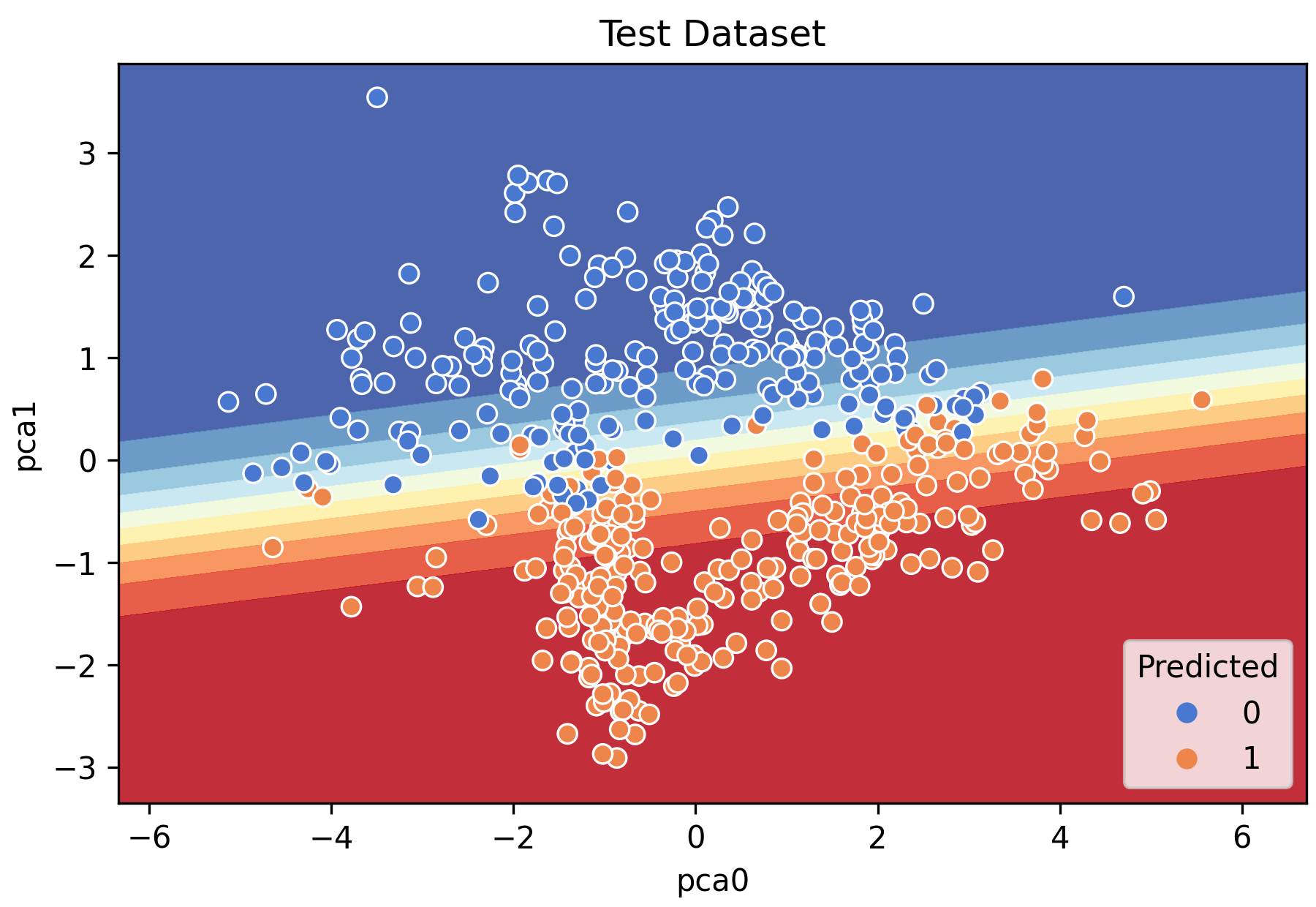}
        \caption{Original distribution}
        \label{subfig:first_round}
    \end{subfigure}
    \hfill
    \begin{subfigure}[t]{0.32\textwidth}
        \centering
        \includegraphics[width=\linewidth]{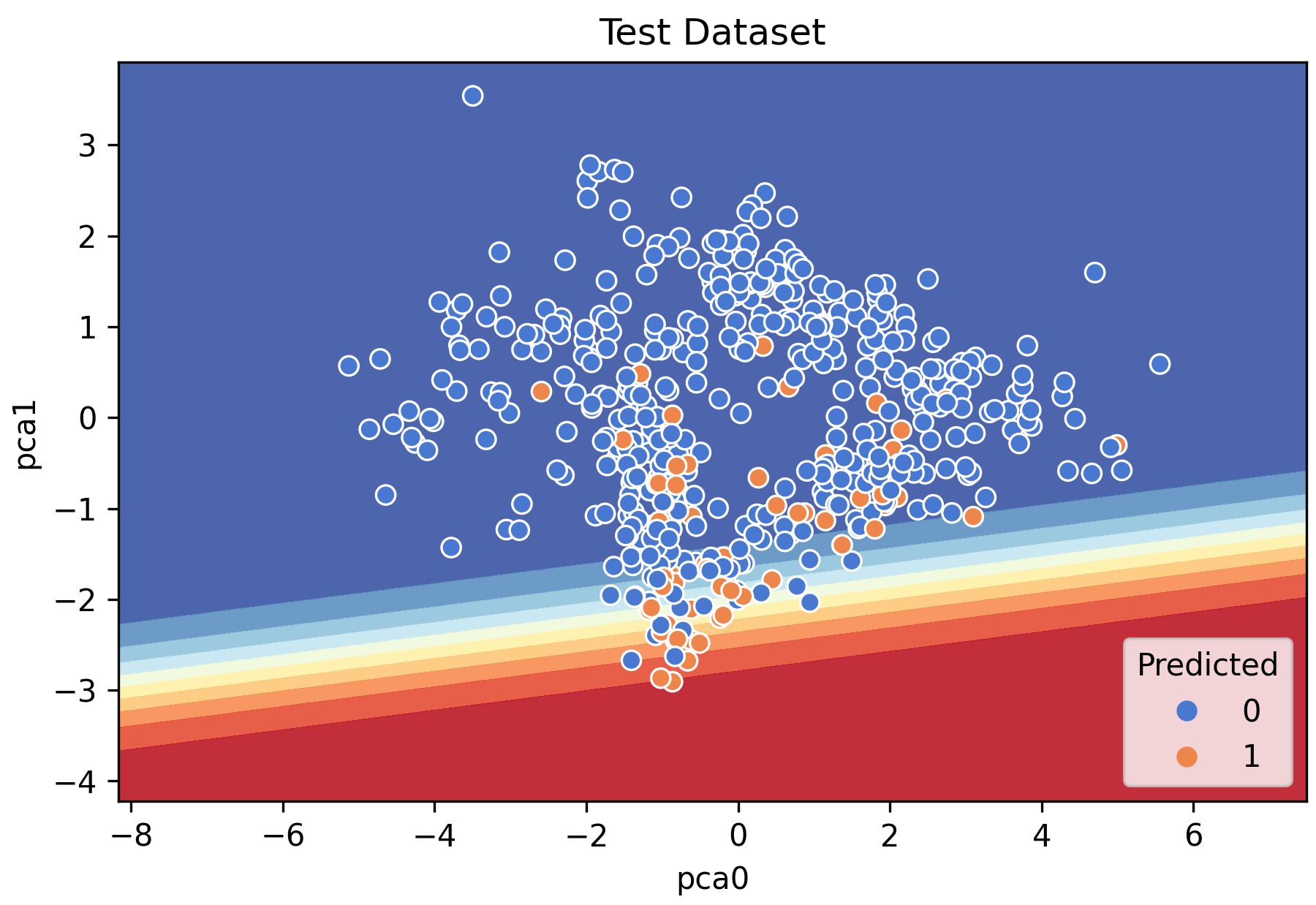}
        \caption{Updated at 30-th. round}
        \label{subfig:middle_round}
    \end{subfigure}
    \hfill
    \begin{subfigure}[t]{0.32\textwidth}
        \centering
        \includegraphics[width=\linewidth]{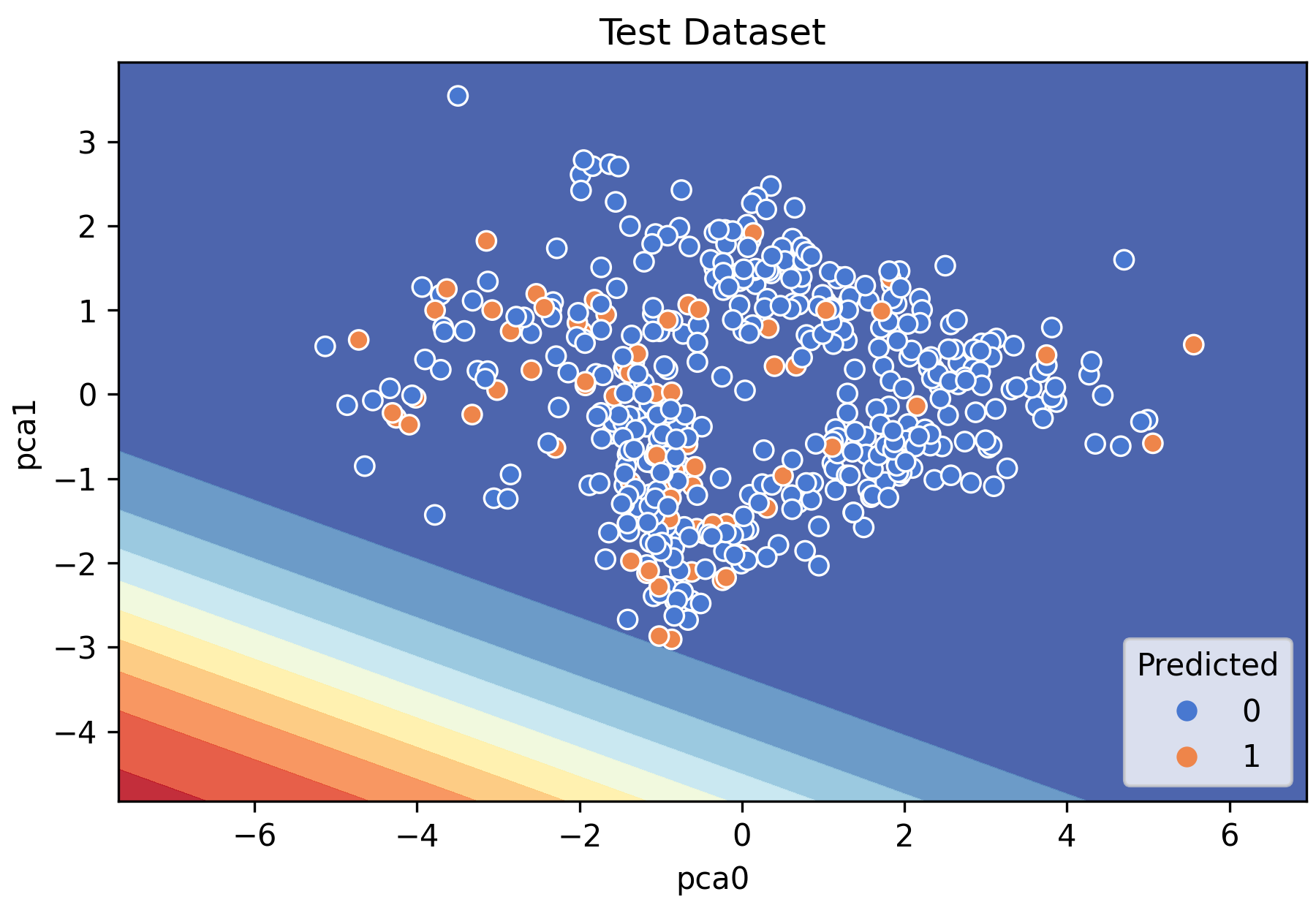}
        \caption{Updated at 70-th. round}
        \label{subfig:last_few_round}
    \end{subfigure}
    
    \caption{The experiment of model evolution with algorithmic recourse, where the model is updated with the top-$k$ labeling. These figures show the test dataset, which is sampled from the original distribution and remains unchanged across all rounds.}
    \label{fig:pre_exp}
\end{figure*}

\paragraph{Experiment:}
Figure~\ref{fig:pre_exp} shows the experimental results with a binary logistic model. The experimental setup is described in Section~\ref{Sec:exp}. Simulations show that the model classifies accepted and rejected data points around half and half during the first few rounds. However, as the number of rounds increases, the decision boundary shifts to a higher standard and rejects majority of users who do not execute recourse actions. 

\section{Related Work}
Algorithmic recourse offers interpretable and actionable guidance to help individuals alter model outcomes, enhancing transparency and accountability in machine learning~\cite{gunning2019xai,arrieta2020explainable,guidotti2018survey}. In high-stakes domains, such as finance~\cite{barocas2017fairness}, healthcare~\cite{bastani2017interpretability,bertossi2020asp,liang2014deep}, and education~\cite{fiok2022explainable,khosravi2022explainable}, recourse mechanisms help individuals understand the reasoning behind model predictions and identify feasible improvement steps.

Most algorithmic recourse methods are designed for one-hop decisions, using gradient-based optimization to identify minimal input changes that flip the model decision~\cite{shao2022gradient}. FACE uses shortest-path methods with density-based metrics~\cite{poyiadzi2020face}, while GDPR emphasizes user-centric explanations that support legal and ethical accountability~\cite{wachter2017counterfactual}.

In a real-world setting, users adapt their behavior in response to model suggestions, influencing future model inputs. Over time, such feedback-driven adaptation can lead to distributional changes. These observations motivate the integration of continual learning into recourse-aware systems, enabling models to adjust dynamically to users' changing behaviors.

\subsection{Data Drift and Continual Learning}

Even without the bi-directional effect on the model and recourse users, it is naturally to have concept shift or data drift when deploying the model over time. There are several work address this issue for the algorithms generating recourse actions. Gao et al. proposed a generative algorithm to mitigate social segregation risks in large-scale algorithmic recourse~\cite{gao2023impact}. Gupta et al. addressed fairness under data shift by equalizing actionable recourse across demographic groups~\cite{gupta2019equalizing}. Other work such as Rawal et al.~\cite{rawal2020wild},  Upadhyay et al.~\cite{upadhyay2021minimizing}, and De Toni et al.~\cite{de2024time} address the reliability of the recourse actions under the concept drifts over time.

From the deployed model side, Continual learning (CL) is particularly relevant for adapting to different distributional dynamics because it allows models to learn from a stream of tasks without catastrophic forgetting~\cite{lopez2017gradient,zenke2017continual,wang2024continual}. Common strategies include regularization-based approaches~\cite{kirkpatrick2017overcoming}, replay-based methods that store or generate past samples for rehearsal~\cite{rolnick2019experience,rebuffi2017icarl}, and dynamic architecture approaches that expand the model structure to accommodate new tasks~\cite{rusu2016progressive,yoon2018lifelong}.

However, these approaches assume that task sequences are externally defined and overlook evolving data distributions. In contrast, our setting is behavior-driven: data adapt to model outputs, leading to feedback-driven distributional changes, which blurs the boundary between continual learning and data shift.

\subsection{Model Deployment with Recourse}

Beyond distributional concerns, recent studies highlight the need for robust algorithmic recourse under model shifts and real-world deployment dynamics. Proposed approaches focus on ensuring recourse coverage, stability under distributional changes, and resilience to feedback loops that may degrade model performance over time~\cite{bui2023coverage, upadhyay2021towards, rawal2023causal}. Other work focus on the algorithms of generating recourse action, which tends to incentives users to perform truthful improvement rather than deceiving the system~\cite{chen2023learning, estornell2023incentivizing}. Unlike designing new recourse algorithms, some work observe and discuss the influence the of recourse users to the deployed model such as increasing classification errors~\cite{fokkema2024risks} or affecting reliability and fairness over time~\cite{bell2024game}. Altmeyer et al. examined evolving models and data shifts in the recourse setting and proposed mitigation strategies~\cite{altmeyer2023endogenous}.

To our knowledge, none of these works investigates the long-term bidirectional effects and properties triggered by the interactions between the system and users. Our proposed research builds on these simulation-based studies and extends them to explore their long-term impacts on both users and systems. We design strategies to mitigate these effects, and foster positive feedback loops benefiting both sides.

\section{Framework}
\label{Sec:framework}

We discuss the decision-making scenario with a set of users and a binary classification model in multiple rounds. Each user is a data point with $d$ features and the model is a score function between 0 and 1 (0 means ``rejected'' and 1 means ``accepted''). In round $t$, a dataset $\mathcal{D}^t$ with size $N$ is sampled from a distribution $P_{data}$, which is a mixed distribution between $\mathcal{D}^{t-1}$ and $\mathcal{D}^{0}$ (i.e., the original distribution). Then, $\mathcal{D}^t$ is modified to $\mathcal{D'}^t$ in which some of the rejected users in $\mathcal{D}^t$ perform recourse actions to modify their features and improve their scores in $h^t$, hoping to flip the results. The system receives the (responded) data set $\mathcal{D'}^t$ and labeled it based on the scores of $h^t$ and the constrained resource $k$. Lastly, the model is updated to $h^{t+1}$ based on $\mathcal{D'}^t$ with the labels.  
The procedure of round $ t $ is shown in the following.

\begin{enumerate}
    \item \textbf{Sample the dataset:}
\[
\mathcal{D}^t = \{ x^{(1)}, x^{(2)}, \dots, x^{(N)} \}, \quad \text{with } x^{(i)} \sim P_{\text{data}}.
\]
  
    \item \textbf{Randomly select rejected users and perform recourse action:}
    
    Randomly select a subset of rejected users $\mathcal{S} \subset \mathcal{D}^t$  and $\forall x \in \mathcal{S}, h^t(x)<0.5$. Modify the features of each user $x$ using recourse function $ r $. The modified set $\mathcal{S}'$ is:
    \[
    \mathcal{S}' = \{r(x) \mid x \in S\}.
    \]
    The updated dataset $ \mathcal{D'}^t $ is:
    \[
    \mathcal{D'}^t = (\mathcal{D}^t \setminus \mathcal{S}) \cup \mathcal{S}'.
    \]

    \item \textbf{Determine new labels for $\mathcal{D'}^t$ with resource $k$:}
    For each user $x \in \mathcal{D'}^t$, the label is determined by the labeling function $f$, which converts the predicted score $ h^t(x)$ into a value in $ [0,1] $ with the hard constraint that the number of accepted samples in $\mathcal{D'}^t$ is at most $k$. 
    That is, for each $x \in \mathcal{D'}^t,$  $f \circ h^t(x) \in [0,1]$ and 
\[
\sum_{x \in  \mathcal{D'}^t} \mathds{1} [f \circ h^t(x) = 1] \leq k
\]

    \item \textbf{Update the model:}
    Use the modified dataset $ \mathcal{D'}^t$ and the new binary labels $\{y^{(i)}\}$, where $y^{(i)}=f \circ h^t(x^{(i)})$, to update the model by minimizing the loss function $ \mathcal{L} $:
    \[
    h^{t+1} = \arg\min_h \sum_{x^{(i)} \in \mathcal{D'}^t} \mathcal{L}(h(x^{(i)}), y^{(i)}).
    \]
\end{enumerate}

This iterative process allows us to explore how recourse actions influence the system dynamics, user behavior, and the evolution of the recommendation model over time. We are specifically interested in Step 2, 3, 4, which we sequentially denoted by \textit{User Response phase}, \textit{Labeling phase}, and \textit{Model Update phase}.

\subsection{User Response Phase}
The user response function $r$ on model $h^t$ can be understood via the recourse action analysis~\cite{verma2020counterfactual,upadhyay2021towards}. The recourse action $x'$ for a user with features $x$ is determined by solving the following optimization problem:
\begin{align*}
x' & = \arg\min_{x'} c(x, x'),\\
& \text{s.t.} \quad  h^t(x') = 1,
\end{align*}
where $c(x, x')$ represents the cost function associated with the action. However, this equation is generally challenging to solve due to the presence of a hard constraint. To address this, most approaches reformulate the problem by Lagrangian relaxation:
\begin{equation}
\label{eq:recourse_loss}
x' = \arg\min_{x'} \ell(h^t(x'), 1) + \lambda c(x, x'),    
\end{equation}
where $\ell : [0, 1] \times [0, 1] \to \mathbb{R}$ is a differentiable loss function (e.g., binary cross-entropy), ensuring that the gap between $h^t(x')$ and the favorable outcome $1$ is minimized. The parameter $\lambda > 0$ acts as a trade-off between minimizing the loss and the cost of the recourse action. In this framework, the quality of recourse can be controlled by replacing the favorable outcome $1$ with some constant $p$ less than 1. Usually, $p$ should be larger than 0.5 to ensure the recourse feature $x'$ can be at least accepted in $h^t$. For example, one typical setting is to consider the $L_2$ distance with constant coefficients (i.e., $c$ is a vector with all positive values).
\begin{equation}
\label{eq:recourse_loss_linear}
x' = \arg\min_{x'} \ell(h^t(x'), 1) + \lambda \sum c_i(x_i'-x_i)^2+c_0  
\end{equation}

\subsection{Labeling Phase}
Given the constrained resource $k$ and predicted score $h(x)$, a classical policy is to apply the top-$ k $ function~\cite{fonseca2023setting}. In this case, the function $ f $ sets the largest $ k $ values of $ h(x)$ to $ 1 $ and the rest to $ 0 $:
    \[
    y = f \circ h(x)= 
    \begin{cases} 
    1 & \text{if } h(x) \text{ is among the top } k \text{ values} \\
    0 & \text{otherwise}
    \end{cases}
    \]

\subsection{Model Update Phase}
Here we consider two settings, a typical setting and continual learning setting. Both settings use ADAM~\cite{kingma2014adam} as the optimizer, where the typical setting uses the cross-entropy loss. The continual learning uses its specific loss function mentioned in Section~\ref{subsec:CLandDCL}.

\section{Analysis on Logistic Model Evolution}
\label{sec: theoretic analysis}

We investigate the influence of recourse users on model evolution. Specifically, we aim to understand how recourse actions contribute to the internal momentum that shapes the direction of model updates over time. Our analysis focuses on the logistic regression model, which serves as a fundamental building block for understanding more complex deep learning architectures.

\begin{definition}[Resource saturation]
    Follow the framework in Section~\ref{Sec:framework}, the system is said to reach \textbf{resource saturation} at round $t$ if the accepted users in $D'$ is exactly $k$.
\end{definition}

In the subsequent analysis, we assume that the system remains fully saturated at all time steps.

\begin{definition}[Higher standard]
\label{definition2}
Given two binary classifiers $h,h'$ and a data distribution $\mathcal{D}$, we say that model $h$ has a \textbf{higher standard} than model $h'$ if and only if:
\[
\mathbb{E}_{x \sim \mathcal{D}}[ h(x) ] < \mathbb{E}_{x \sim \mathcal{D}}[ h'(x) ].
\]
\end{definition}

\begin{theorem}[Higher Standard Provides Higher Accuracy]
\label{thm: increase boundary}
Let $D,h$ be a dataset and the logistic model respectively at round $t$. For the responded dataset $D'$, there exists a model $h'$ that has a higher standard than $h$ and achieves higher accuracy on $D'$ if and only if there exists at least one recourse user, newly labeled as positive (i.e., class 1), whose score ranks within the top-$k$ in $D'$. 
\end{theorem}

The proof is provided in the Appendix. In summary, if a recourse user enters the top-$k$ ranked scores in $D'$, the score threshold corresponding to the $k$-th user in $D'$ is higher than that in the original dataset $D$. As a result, the decision boundary of $h$ can be shifted to this higher threshold, thereby rejecting all users in $D'$ with scores below the $k$th ranked user. This effectively increases the model's standard (i.e., makes it stricter) and results in improved classification accuracy over $D'$.

While Theorem~\ref{thm: increase boundary} illustrates that pursuing higher accuracy may lead to a more strict decision standard, accuracy is not the typical optimization objective in most learning settings. Therefore, we turn our attention to cross-entropy loss (CE loss), the standard objective in logistic regression, and examine how recourse behavior influences model evolution under this loss.

\begin{lemma}[Recourse action]
\label{lemma: recourse action}
Given a logistic model $h$ and a $L_2$ cost function $c$ (Equation~\ref{eq:recourse_loss_linear}), the recourse action taken by a user on feature $i$ is proportional to $\frac{w_i}{c_i}$, where $w_i$ and $c_i$ denote the weights associated with feature $i$ in the model and the cost function, respectively.

\end{lemma}
\begin{proof}
    Denote $h(x)=1/(1+e^{-(wx+b)})$. Referring to Equation~\ref{eq:recourse_loss_linear}, the cost of action $a$ on recourse function $r$ is in the following.
    $$
    r(a)=-(\log h(x+a))+\lambda\sum c_i a_i^2+c_0.
    $$
    Since $r$ is convex, we can calculate the optimal action $a$ via the partial derivative is zero at any dimension $i$. That is,
    $$
    \frac{\partial r}{\partial a_i} = -w_i(1-h(x+a))+2a_ic_i \lambda =0.
    $$
     This implies 
    $$
    a_i=\frac{1-h(x+a)}{2\lambda} \cdot \frac{w_i}{c_i}
    $$
\end{proof}

Lemma~\ref{lemma: recourse action} illustrates that recourse actions are biased in proportion to $\frac{w_i}{c_i}$. Based on this insight, our first observation is that model updates driven by failed recourse users tend to result in a new model $h'$ with a higher standard under the cross-entropy loss. This is formally stated in Proposition~\ref{prop: failed recourse}.

\begin{prop}[Failed Recourse Actions Drive Higher Standards]
\label{prop: failed recourse}
Under the cross-entropy loss, let $D$ be a dataset and $h$ be the logistic model trained on $D$. If a user $x$ performs a recourse action resulting in $\hat{x}$, but is still assigned label 0 in the updated dataset $D'$, then the resulting model $h'$ will have a higher standard than $h$.
\end{prop}

\begin{proof}
    Since $\mathcal{L}_{CE}(h(x),y)$ is convex, the partial gradient of it is zero among all dimensions over $D$. That is, $\forall i \in [1,d]$
    $$
    \sum_{x \in D} \frac{\partial \mathcal{L}_{CE}(h(x),y)}{\partial w_i}=0,
    $$
    which gives
\begin{equation}
\label{eq:two_terms}
\sum_{x \in D^+} x_i(h(x)-1)+\sum_{x \in D^-} x_i(h(x))=0,    
\end{equation}
where $D^+,D^-$ denote the set of users with label $1$ and $0$, respectively. Assume $w_i,x_i>0$. In Equation~\ref{eq:two_terms}, the first term contributes the negative gradient and the second term contributes the positive one, denoted by $\mathcal{H}^+_{D,h}, \mathcal{H}^-_{D,h}$ respectively. One can think the value of $w_i$ is changed in order to balance the two terms, i.e., sum to zero.

Denote the recourse user before and after the recourse action as $x$ and $\hat{x}$. Notice that both $x$ and $\hat{x}$ belong to $D^- ,D'^-$ but $\hat{x}$ has a higher score and feature value in dimension $i$ (via Lemma~\ref{lemma: recourse action}). Thus, the value of $\mathcal{H}^-_{D',h}$ is higher. Now, consider the situation that the updated model $h'$ is found via gradient descent starting from the parameters of $h$. The direction of changes in $w_i$ is to balance $\mathcal{H}^+_{D',h},\mathcal{H}^-_{D',h}$. Since $x_i$ is unchangeable, the adjustment only affects the output score of $x$. That is, after $h'$ fits to $D'$, we have 
$$
\sum_{x \in D^+} x_i(h(x)-1) > \sum_{x \in D'^+} x_i(h'(x)-1)
$$
and 
$$
\sum_{x \in D^-} x_i(h(x)) < \sum_{x \in D'^-} x_i(h'(x)).
$$
The above inequalities show that the updated model $h'$ provides lower score than $h$.
\end{proof}

On the other hand, if the resource is shortened, even if there is no recourse action, the updated model $h'$ also leads to a higher standard.

\begin{prop}[Limiting Resource Drives Higher Standard]
\label{prop: limiting recourse}
Under the cross-entropy loss, let $D,h$ be the dataset and the logistic model fit to $D$. If a user $x$ labeled as 1 in $D$ but 0 in $D'$ due to the limited resource, it drives the updated model $h'$ into a higher standard.
\end{prop}
One can prove Proposition~\ref{prop: limiting recourse} by a similar analysis in Proposition~\ref{prop: failed recourse}, which we skip here. 

However, the system becomes complicated when some recourse users replacing the non-recourse users via top-$k$ policy. To see that, denote by $u$ the user labeled as 1 in $D$ and replaced by some recourse user $\hat{x}$ via top-$k$ policy. With a similar analysis with partial derivative on $i$, we have
$$
H_{D',h}^-=H_{D,h}^- - x_ih(x) + u_i h(u),
$$
and
$$
H_{D',h}^+=H_{D,h}^+ + \hat{x_i}(h(\hat{x})-1) - u_i (h(u)-1).
$$
Reorganizing,
$$
H_{D',h}^-+H_{D',h}^+=u_i-x_ih(x)-\hat{x_i}+\hat{x_i}h(\hat{x}).
$$

Denote the above term by $F(x_i,\hat{x_i},h(x),h(\hat{x}),u_i)$. The positivity of $F$ subject to the following constraints.
\begin{align*}
\text{Subject to:} \quad & 0 \leq x_i<\hat{x_i}  \\
                         & h(x)<h(\hat{x}) \\
                         & 0\leq h(x)<.5\\
                         & .5\leq h(u) \leq 1 \\
                         & .5\leq h(\hat{x}) \leq 1 
\end{align*}
Due to the complexity of $F$, which involves 5 variables and product of variables, we explicit list some conditions that make $F$ positive or negative.
\begin{equation}
F(x_i,\hat{x_i},h(x),h(\hat{x}),u_i)
\begin{cases}
> 0\text{,} &  u_i \geq (1-\alpha)\hat{x_i} \\
> 0\text{,} &  x_i < u_i < \hat{x_i} \text{ and } \frac{x_i}{\hat{x_i}} > \beta \\
< 0\text{,} & u_i \leq x_i \text{ and } \frac{x_i}{\hat{x_i}}< \beta \\
< 0\text{,} & u_i < \frac{1}{2} x_i,
\end{cases}
\end{equation}
where $\alpha= h(\hat{x})-h(x), \beta= \frac{1-h(\hat{x})}{1-h(x)}$. 

Generally, we observe three key patterns. First, if the replaced user has a significantly higher feature value (i.e., $u_i > x_i$), the resulting model $h'$ tends to adopt a higher standard; conversely, if the value is much lower (e.g., $u_i < \frac{1}{2} x_i$), the standard tends to drop. Second, the greater the score improvement from the recourse action, the more likely $h'$ is to shift towards a higher standard. Third, when the change in the feature value $x_i$ is disproportionately large compared to the resulting score change---indicated by a high $\beta$ value---$h'$ tends to adopt a higher standard. The second and third observations suggest that intense and biased recourse behavior can drive the model toward a more strict decision threshold.

\section{Methods Toward Robust Model Evolution}
\label{sec:method}

Section~\ref{subsec:fair_topk} and Section~\ref{subsec:CLandDCL} are our two proposed methods to address the issue of standard shifting.

\subsection{Fair Top-$k$ Policy}
\label{subsec:fair_topk}
To address the issue of standard shifting, the intuition is to select a set of users that are ``diverse'' from each other. That is, if a set of points with high scores is similar to each other, we would decrease the chance of selecting all of them as accepted points. To do so, we use kernel density estimation (KDE) to assess the density of similar data points, which provides a measure of local crowdedness. Next, we generate a biased weight vector $v$ by combining each kernel density score inverse with $\kappa\hat{y}_i$, where $\kappa$ is a dimensionless weighting factor that balances the contribution of $\hat{y}_i$. Using this weighted distribution, we randomly select $k$ data points from those where $f(\hat{y}_i) = 1$. The remaining unselected data points are assigned $y_i = 0$, indicating rejection.
\begin{equation}
    v_i = KDE^{-1}(x_i) + \kappa\hat{y}_i
\end{equation}
However, after the labeling phase, the accepted and rejected data become mixed due to random selection, which can negatively impact model training. To mitigate this issue, we remove some negative data points from the training set if they satisfy $h(x_i) > 0.5$.

\subsection{Continual Learning and DCL}
\label{subsec:CLandDCL}
To prevent model from drastic shifting, we use continual learning as a method to memorize past distributions. Specifically, we adapt Synaptic Intelligence (SI)~\cite{zenke2017continual}, a classical continual learning method to achieve our goal. The original SI is a regularization-based continual learning method, with loss regularization defined as follows:
\begin{equation}
    \mathcal{\tilde{L}}_t = \mathcal{L}_t + \tau\sum_k\Omega^t_k(\tilde\theta_k - \theta_k)^2
    \label{surrogate loss SI}
\end{equation}

\begin{equation}
    \Omega^t_k = \sum_{u<t}\frac{\omega^u_k}{(\Delta^u_k)^2 + \epsilon}
    \label{omega SI}
\end{equation}
In equation \eqref{surrogate loss SI} , $\tau$ is a dimensionless scaling factor that regulates the contribution of the previous task's weight. $\theta_k$ represents each individual parameter of the current parameter set and $\tilde{\theta}_k = \theta_k(t-1)$. $\Omega^t_k$ is the per-parameter regularization strength. In equation \eqref{omega SI} $, \omega^u_k$ is the per-parameter loss in each task while $\Delta^u_k = \theta_k(u)- \theta_k(u-1)$. $\epsilon$ is a small value to prevent zero-division.

In our scenario, we are not interested in the loss among all past distributions but only a few past-rounds. To focus on short-term tasks rather than all tasks before task \( t \), we modify the function \( \Omega^t_k \) by introducing a learning range \( r \) and a dimensionless weight constant \( w_u \). The weight \( w_u \) follows a cubic distribution, assigning greater importance to tasks closer to \( t \).
\begin{equation}
    \Omega^t_k = \sum^{t-1}_{u = t-r}w_u\frac{\omega^u_k}{(\Delta^u_k)^2 + \epsilon}
\end{equation}
The previous continual learning setting set constant value $\tau$ among all rounds, which is not flexible when facing different data distributions. To address this, we introduce \textbf{Dynamic Continual Learning (DCL)} by modifying $\tau$ to  
\begin{equation}  
    \tau_t = \frac{\tau}{JSD_{t-1}}  
\end{equation}

Here we use the Jensen-Shannon Divergence (JSD) $\in [0, 1]$~\cite{jensen1998divergence} as an evaluation metric to measure the distance between the positive and negative data distributions. It allows the strength of past tasks to be adjusted based on the previous round’s level of chaos. Hence, if the last task is nearly collapsed, $JSD_{t-1}$ decreases, leading to an increase in $\tau_t$. This prevents the model from aggressively learning new patterns that could introduce further instability into the model.

\section{Experiments and Observations}
\label{Sec:exp}

The experiments aim to simulate a virtual scenario where the decision model that
incorporates a recourse function over the long term, starting with training on a fixed distribution of data points. Our experimental setup follows the framework introduced in Section~\ref{Sec:framework}. We use classical two-layers MLP and logistic regression as the decision models on three different datasets. The resource constraint $k$ is set to $\frac{N}{2}$. The labeling phase includes Top-$k$ and our proposed method: Fair-top-$k$. In the model update phase, we use ADAM with Binary Cross Entropy (BCE) loss, continual learning loss, and dynamic continual learning (DCL) loss we described in Section~\ref{subsec:CLandDCL}. Both The constant $\tau$ used is $10^{-5}$ in DCL and typical continual learning. The hyperparameter $\kappa$ for Fair-top-$k$ is set as $10^{-4}$.

We use both synthetic and two real-world data to simulate our virtual environment. The synthetic dataset is generated in $\mathbb{R}^{20}$, with 17 dimensions that are all actionable (mutable).~\textit{UCI defaultCredit} dataset~\cite{yeh2009comparisons} is related to customers' default payments and has 23 features with 19 of them are actionable. In ~\textit{Credit} dataset we referred the work from Ustun et al.~\cite{ustun2019actionable} and set 11 actionable features.
 
During the recourse phase, Equation~\ref{eq:recourse_loss} is used to generate the recourse action. The cost function is calculated among all mutable features. Denote $x=<x_1,\cdots,x_d>$ and $x'=<x_1',\cdots,x_d'>$ 
\[
c(x', x) = \sum_{i, i \in mutable } c_i\left( x_i' - x_i \right)^2 + c_0.
\]  
We set $c_0$ to 0 so there is no cost if no action is taken. In real-world datasets, the weights $\{c_i\}$ are set to reflect the real-world scenarios. In the synthetic dataset, we try different weight distributions including uniform, inverse gamma distribution, and logarithm distribution.

\subsection{Metrics}

We use several metrics to evaluate the model stability, model robustness, and recourse fairness of the interaction between system and users.

\subsubsection{Short-Term Balanced Accuracy}
We adapt the concept of Average Accuracy (AA)~\cite{RiemannianWalkAA,GradientAA} and modify it to the evaluation focuses on recent performance. Let \( a_{t,j} \in [0, 1] \) represent the classification accuracy of the \( j \)-th round task on the \( t \)-th round model.
\begin{equation}
    AA_t = \frac{1}{t}\sum^t_{j = 1}a_{t,j}
\end{equation}

Given that data will become imbalanced over rounds, negatively impacting prediction accuracy, we replace $a_{t,j}$ with balanced accuracy $b_{t, j}$~\cite{BalancedAcc}. Balanced accuracy is calculated as the average of recall and specificity, making it a more suitable metric for imbalanced datasets. Here, the balanced accuracy $b_{t,j}$ is defined as:
\begin{equation}
    b_{t,j} = \frac{1}{2}(\frac{TP_{t,j}}{TP_{t,j}+FN_{t,j}} + \frac{TN_{t,j}}{TN_{t,j}+FP_{t,j}})
\end{equation}

To capture the short-term focus, we introduce \(r\) to the equation, defining the number of past rounds considered in the calculation.

\begin{equation}
        {STBA}_t = \sum^{t-1}_{j = t-r} b_{t,j}
\end{equation}
Additionally, we exclude the current round from the calculation to ensure an unbiased evaluation of the model's true performance, as the model is trained on the \( t \)-th task.

\subsubsection{Higher Standard}
The details of this metric are explained in Definition~\ref{definition2}. Here, $\mathcal{D}$ represents the test dataset, and we track the value of the expected output logit before the sigmoid layer: $\mathbb{E}_{x \sim \mathcal{D}}[h'(x)]$. Notably, due to the sigmoid function's asymptotic nature, its output is not ideal for observation (values become very close in each round). Therefore, we analyze $h'(x)$, the model output logit prior to the sigmoid transformation. Furthermore, since the sigmoid function is monotonically increasing, this change does not affect the inequality relationships among higher standard values.

\subsubsection{Average Recourse Cost}
For each recourse user, their recourse cost is $\sum_i c_i(x_i'-x_i)^2 + c_0$, as described in Equation~\ref{eq:recourse_loss_linear}. We set $c_0$ to 0 because there is no cost if no action was taken. This metric calculates the average recourse cost of each recourse user. 

\subsubsection{Fail to Recourse (FTR)}
Users care about the effectiveness of their recourse actions. While ensuring stability, system also needs to maintain fairness for recoursed users. Hence, we introduce Fail to Recourse (FTR) to quantify the overall effectiveness of recourse action, regarding the system. It calculates the proportion of recoursed data in round $t$ that fail to be classified as positive in $t+1$ round. While Fonseca et al.~\cite{fonseca2023setting} 
proposed recourse reliability (RR) as the proportion of recoursed data in round $t$ that classified as positive in $t+1$ round, we do it the opposite way. $R_t$ denotes the subset of all recoursed data and $N_{t+1}$ denotes the subset of classification result which is negative after $t+1$ round of training.
\begin{equation}
    {FTR}_t = \frac{|R_t \cap N_{t+1}|}{|R_t|}
\end{equation}

\subsubsection{Test-Acceptance Rate}
 
To quantify the increasing decision boundary, we introduce the Test-Acceptance Rate (TAR) as an indicator to observe the model's classification standards. Specifically, we examine the ratio of labels \( 1 \) and \( 0 \) at $t$-th round in the test dataset, which reflects the original distribution of the test data, to understand how the model's classification criteria change. Here $T^1$ denotes the data with label $1$ and $T^0$ denotes the data with label $0$ in the test data.
\begin{equation}
    {TAR}_t = \frac{{|T^1}_t|}{{|T^0}_t|}
\end{equation}

\subsection{Top-$k$ Labeling with Static Learning}
\begin{figure*}[t]
    \centering
    \includegraphics[width=0.3\linewidth]{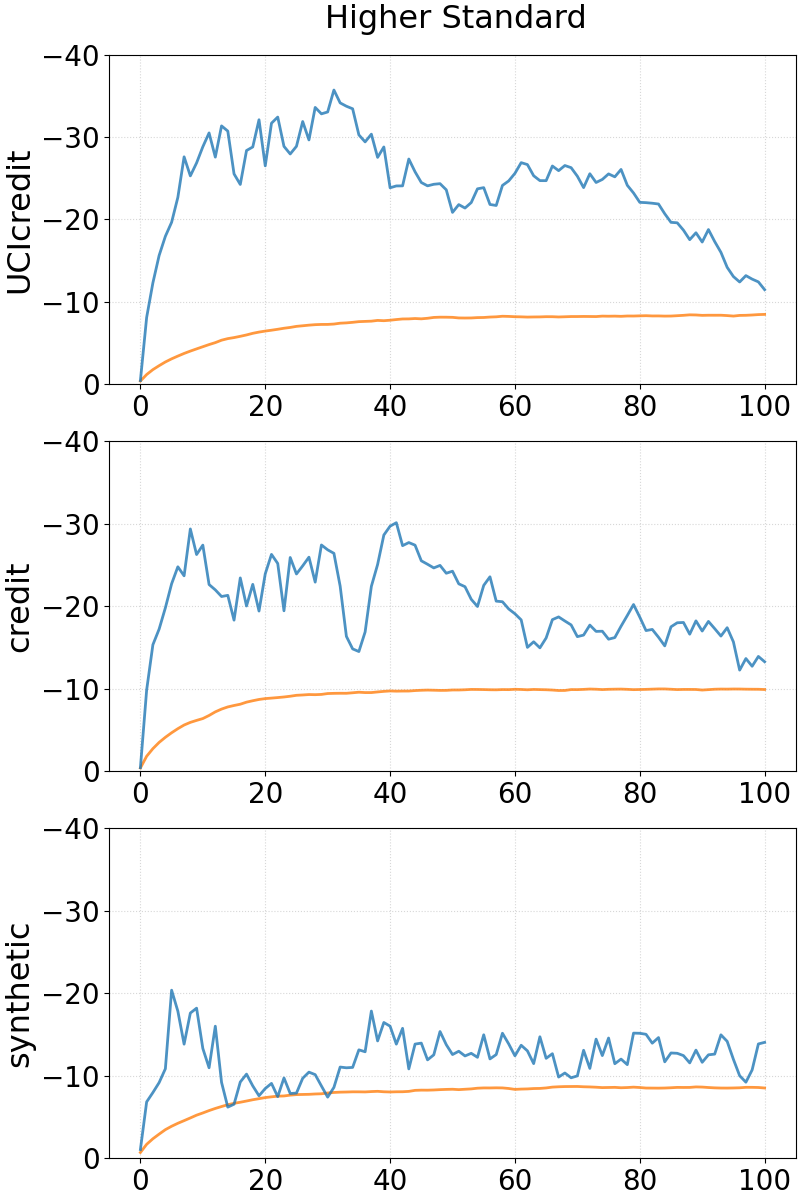}
    \includegraphics[width=0.3\linewidth]{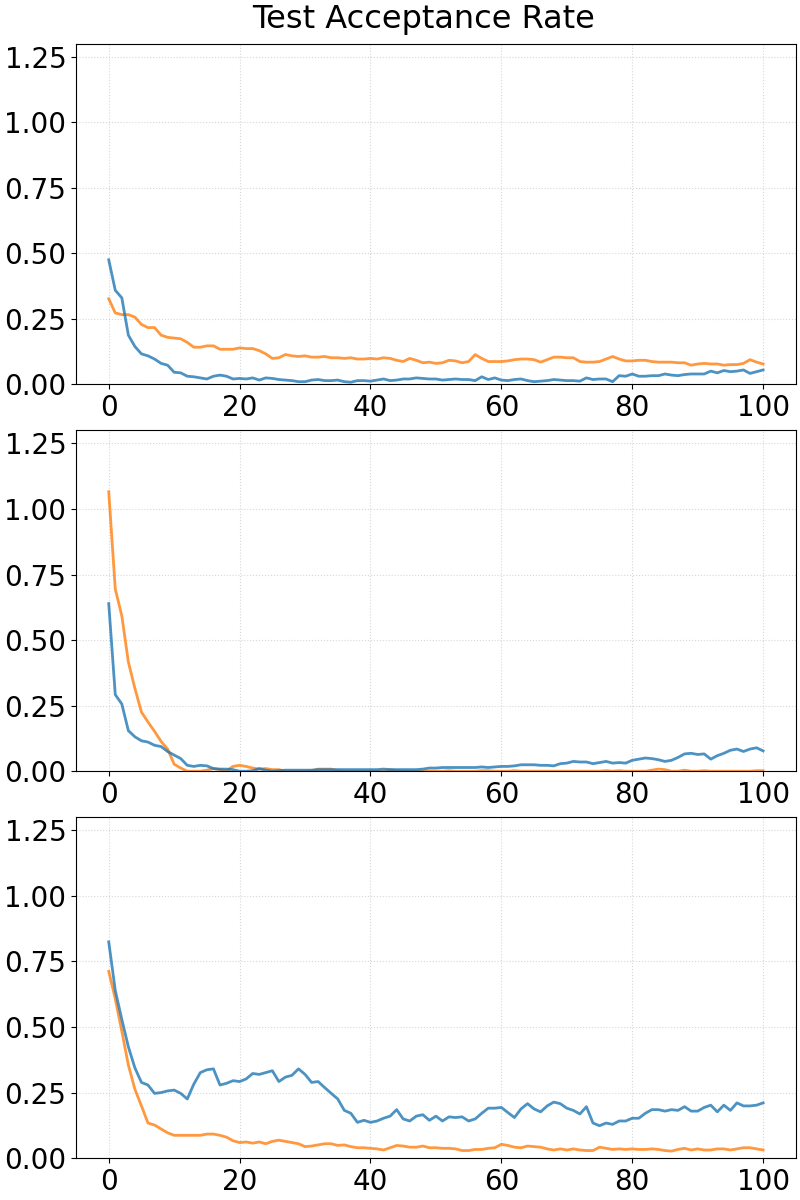}
    \includegraphics[width=0.3\linewidth]{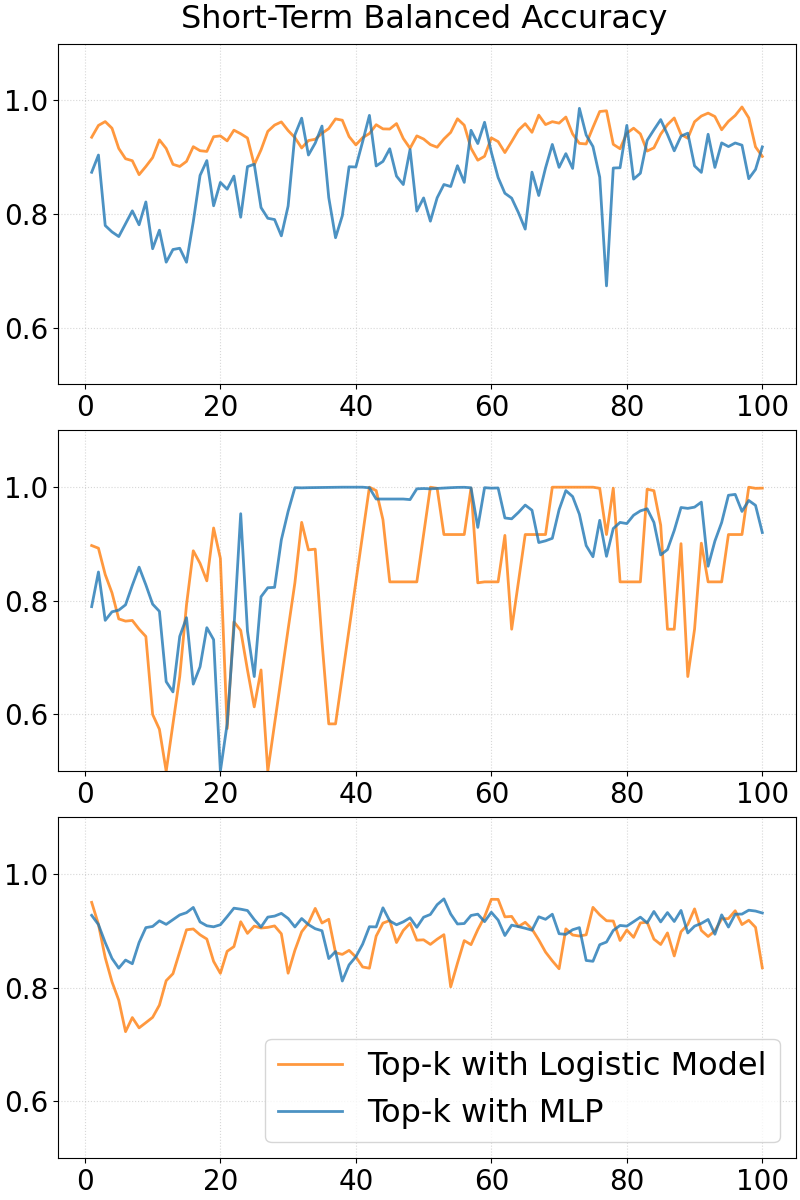}
    \caption{Higher Standard, Test Acceptance Rate, and STBA on Logistic Regression Model and MLP across three datasets.}
    \label{fig:topk and normal update}
\end{figure*}

\begin{figure}[ht]
    \centering
    \begin{subfigure}[b]{0.49\linewidth}
        \centering
        \includegraphics[width=\linewidth]{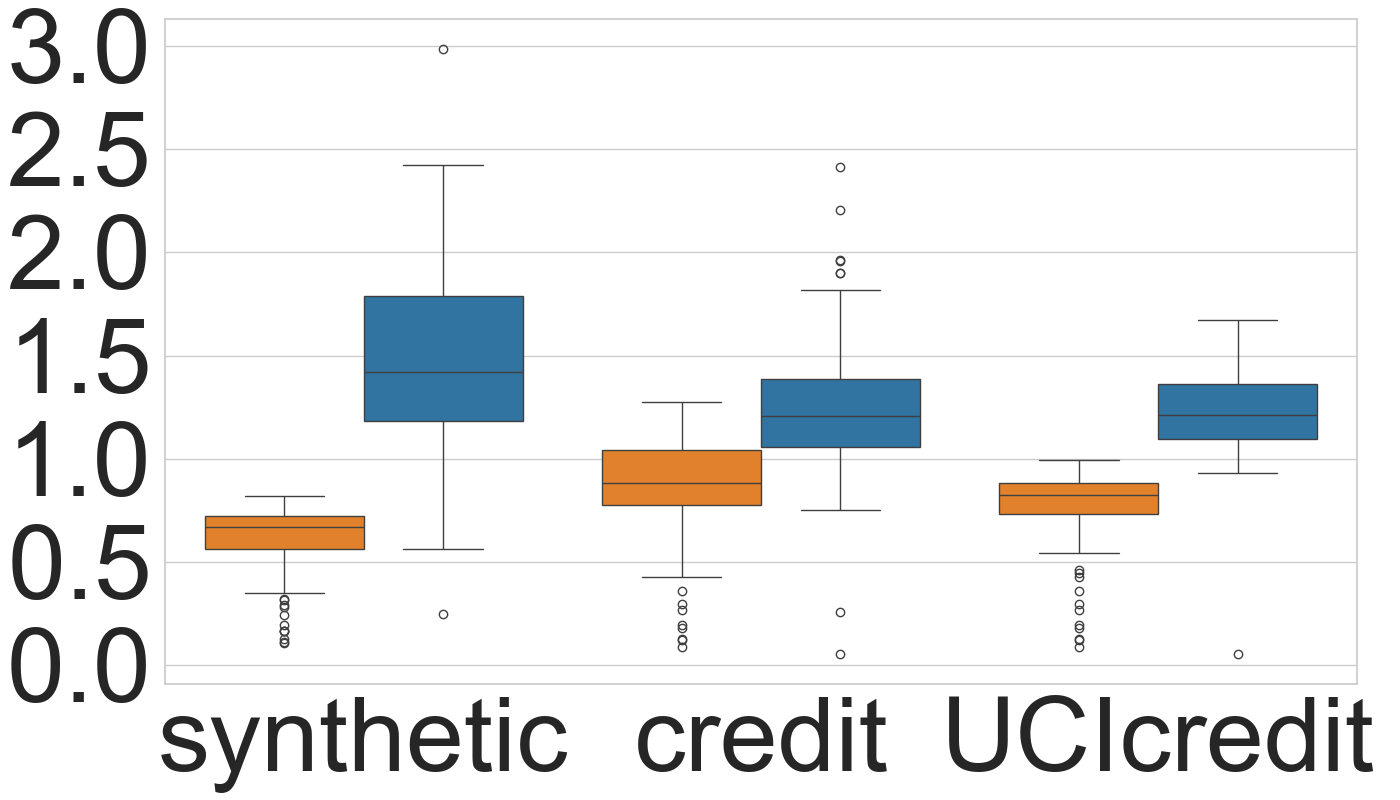}
        \caption{average recourse cost}
        \label{fig:avg_recousre_cost_metrics on normal}
    \end{subfigure}
    \hfill
    \begin{subfigure}[b]{0.49\linewidth}
        \centering
        \includegraphics[width=\linewidth]{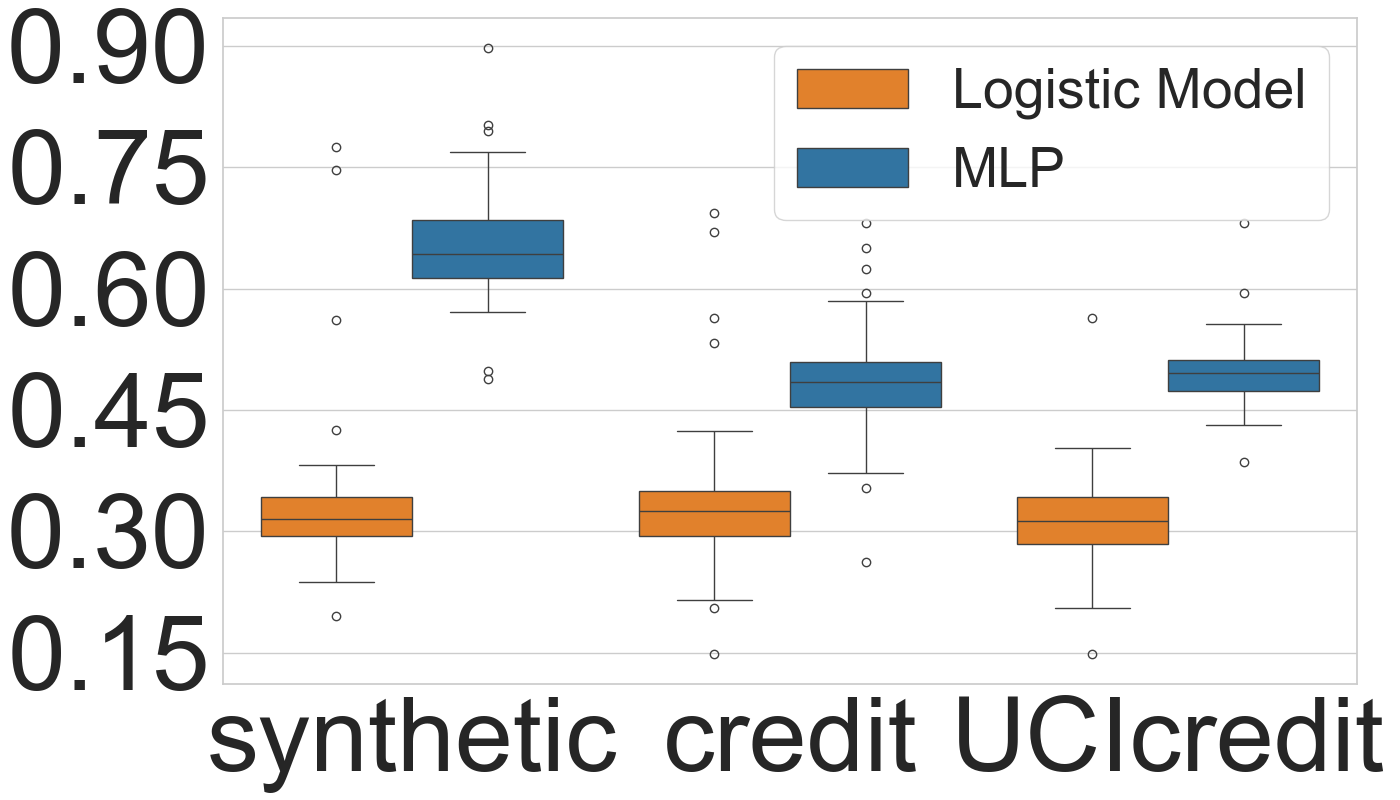}
        \caption{FTR}
        \label{fig:FtR_metrics on normal}
    \end{subfigure}
    \caption{(a) average recourse cost of Top-$k$ method across three datasets. (b) Fail to Recourse (FTR) of Top-$k$ method across three datasets.}
    \label{fig:merged_roe_ftr}
\end{figure}

Figure~\ref{fig:topk and normal update} shows the simulation result in the static learning when the top-$k$ policy is used in labeling phase and cross-entropy loss is set in model update phase. The x-axis is the number of rounds, and the y-axis is the value of the measured metric, which from left to right are Higher Standard, Test Acceptance rate, and Short-term Balanced Accuracy. Each row represents one of the datasets. We have three observations.

 \textbf{Higher Standard occurred.} As illustrated in Figure ~\ref{fig:topk and normal update}, the higher standard metric shows that the model outputs lower score, leading to stricter acceptance criteria across the three datasets, as evidenced by the decreasing test acceptance rates. The test acceptance rates across three datasets are all close to zero after a few rounds, where the value is generally greater than 0.5 in the initial distribution. The highest value is around 0.25 after a few rounds, shown from MLP model in synthetic dataset. This indicates that there are less than 20\% samples that will be accepted after several rounds if they are from the initial distribution.

\textbf{High recourse cost and a high failure rate are observed in MLP models.}
Figure~\ref{fig:avg_recousre_cost_metrics on normal} presents the average recourse cost for both the Logistic and MLP models. The MLP model yields a higher recourse cost compared to the Logistic model. The MLP model is more flexible to the new data distribution, so the model tends to shift more, therefore requiring users to exert more effort to satisfy its criteria, leading to higher recourse cost. Furthermore, the inherent flexibility of MLP model can also introduce instability issues. According to Figure~\ref{fig:FtR_metrics on normal}, the MLP model's Fail to Recourse is markedly higher than that of the logistic model. As indicated by the higher standard metric in Figure~\ref{fig:topk and normal update}, the drastic changes in the MLP's output logits underscore the model instability and the difficulty in aligning recourse.

\textbf{Non-robust accuracy is observed in both models.} According to Figure~\ref{fig:topk and normal update}, the Short-Term Balanced Accuracy is notably low for the \textit{Credit} dataset. This can be attributed to the data distribution of \textit{Credit}, where positive samples in the \textit{Credit} dataset are closer to the decision boundary. Consequently, the test acceptance rate for positive instances drops significantly, leading to only a few positive samples existing in the test dataset. Therefore, the balanced accuracy calculation is heavily penalized if the classification of these positive instances changes between rounds. This leads to a great fluctuation on all Short-Term Balanced Accuracies.

The above observations are also consistent across varying levels of recourse quality (from 0.7 to 1.0) and different ratios of recourse users (from 0.2 to 0.7). While other factors may also influence these patterns, due to space constraints, we provide a summary in Table~\ref{tab:top-k} and include the full experimental details in the Appendix. Overall, increasing the recourse quality and the proportion of recourse users tends to raise the decision standard and reduce the robustness of the model’s predictions.

\setlength{\tabcolsep}{1mm}

\begin{table}[ht]
\centering
\fontsize{9pt}{11pt}\selectfont

\begin{tabular}{|c|c|c|c|c|}

\hline

\diagbox[width=2.6cm, height=1.0cm, font=\normalsize\bfseries]{Setting}{Situation}
& \makecell{Non-robust\\model} 
& \makecell{Recourse\\cost} 
& \makecell{Recourse\\fail rate}
& \makecell{Higher\\standard}  \\
\hline
\makecell{Higher recourse \\ quality} & $\uparrow$ & $\uparrow$ & $\downarrow$ & $\uparrow$ \\
\hline
\makecell{More recourse \\ users} & $\uparrow$ & $\uparrow$ & $\uparrow$ & $\uparrow$ \\
\hline
\makecell{Larger $k$ values} & $\downarrow$ & -- & -- & $\downarrow$ \\
\hline
\makecell{Inverse gamma \\ cost weights} & -- & $\downarrow$ & -- & -- \\
\hline
\makecell{Higher ratio of \\ original distri.} & $\uparrow$ & $\downarrow$ & $\downarrow$ & $\uparrow$ \\
\hline
\makecell{From logistics \\ to MLP} & -- & $\uparrow$ & $\uparrow$ & $\uparrow$ \\
\hline
\makecell{From logistics \\ to conti. learning} & -- & $\downarrow$ & -- & -- \\
\hline
\end{tabular}
\caption{Effect of Different Settings on top-$k$ Method.}
\label{tab:top-k}
\end{table}

\subsection{Fair-top-$k$, Continual Learning, and DCL}

\begin{figure*}
    \centering
    \includegraphics[width=0.3\linewidth]{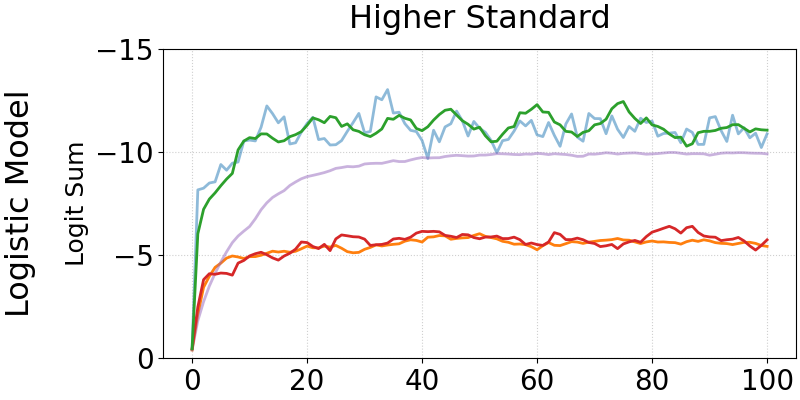}
    \includegraphics[width=0.3\linewidth]{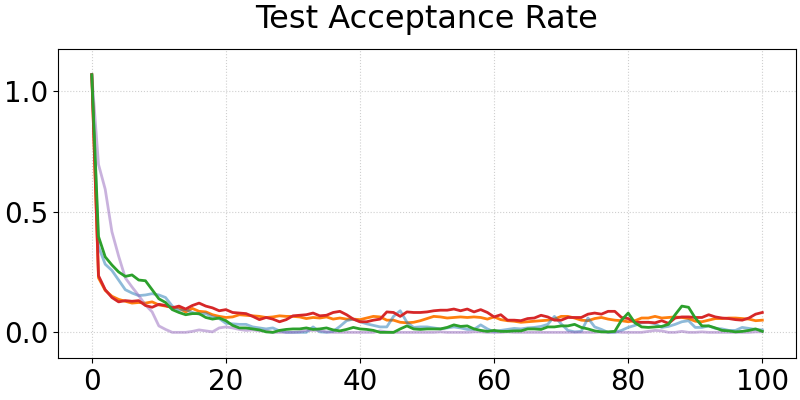}
    \includegraphics[width=0.3\linewidth]{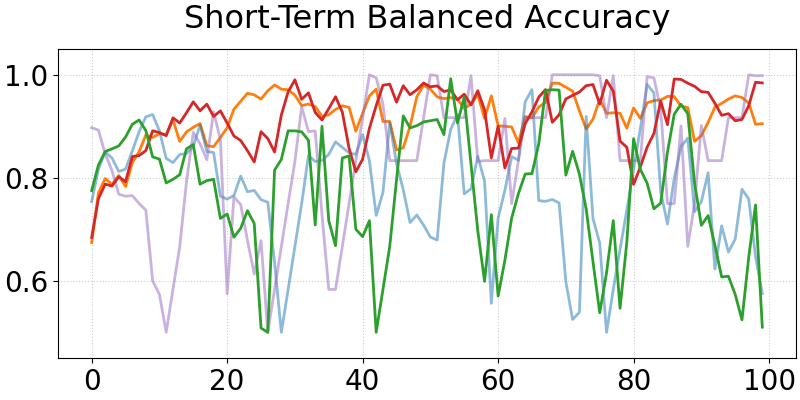}
    \includegraphics[width=0.3\linewidth]{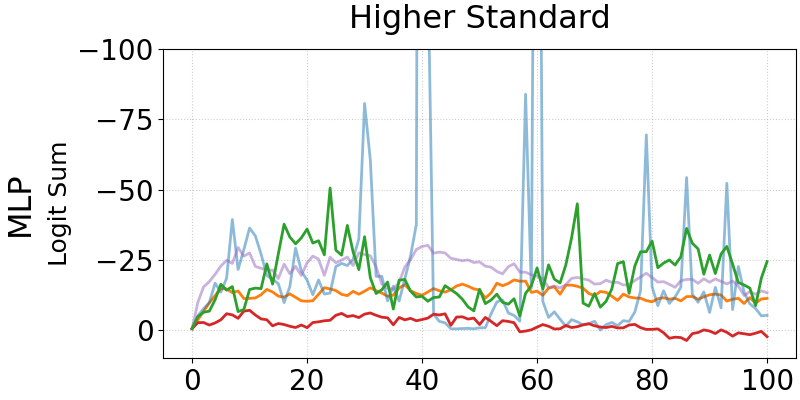}
    \includegraphics[width=0.3\linewidth]{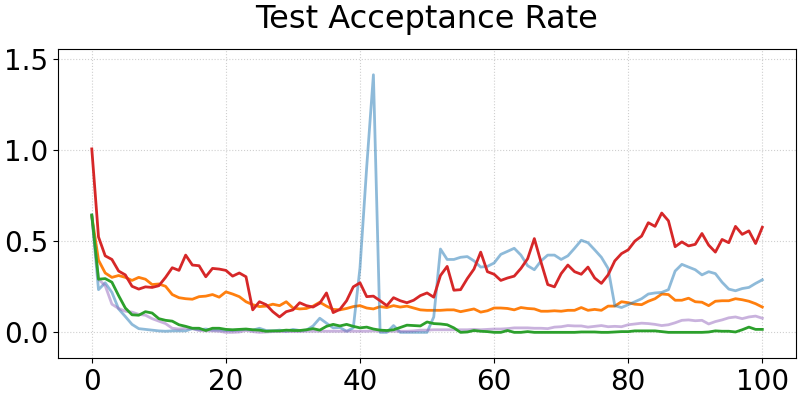}
    \includegraphics[width=0.3\linewidth]{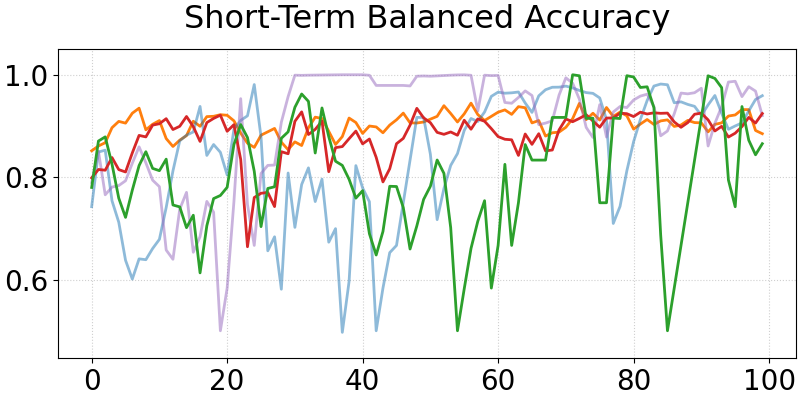}
    \includegraphics[width=0.7\linewidth]{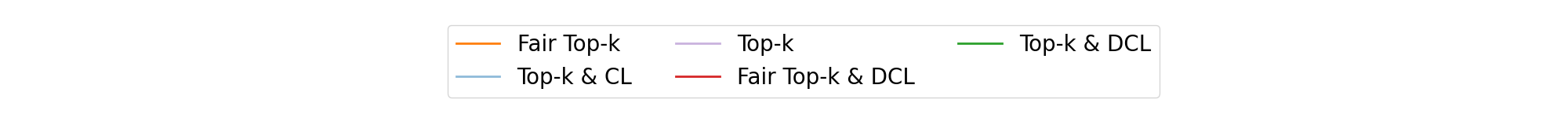}
    \caption{Higher Standard, Test Acceptance Rate, and STBA on Logistic Regression Model and MLP on \textit{Credit} data. The outlier of Higher Standard on MLP with Top-$k$ \& CL method is -776.66, -132.24 at round 40, 41 and -360.13 at round 60.}
    \label{fig:comparison: ours and baselines}
\end{figure*}

\begin{table*}[h!]
     \centering
     \begin{tabular}{ll*{6}{c}}
         \toprule
        \multirow{2}{*}{\textbf{Dataset}} & \multirow{2}{*}{\textbf{Methods}} & \multicolumn{2}{c}{\textbf{Average Recourse Cost}} & \multicolumn{2}{c}{\textbf{Fail to Recourse}} & \multicolumn{2}{c}{\textbf{Higher Standard}} \\
         \cmidrule(lr){3-4} \cmidrule(lr){5-6} \cmidrule(lr){7-8}
         & & \textbf{Logistic} & \textbf{MLP} & \textbf{Logistic} & \textbf{MLP} & \textbf{Logistic} & \textbf{MLP} \\
         \midrule
         \multirow{5}{*}{Synthetic} & Top-$k$ & $0.60 \pm 0.02$ & $1.46 \pm 0.05$ & $\textbf{0.32} \bm{\pm} \textbf{0.01}$ & $0.65 \pm 0.01$ & $-7.64 \pm 0.16$ & $-12.05 \pm 0.30$ \\
         & Top-$k$ \& CL & $0.25 \pm 0.01$ & $0.60 \pm 0.03$ & $0.33 \pm 0.01$ & $0.70 \pm 0.01$ & $-8.33 \pm 0.10$ & $-4.50 \pm 0.50$ \\
        & Top-$k$ \& DCL & $0.38 \pm 0.01$ & $\textbf{0.55} \bm{\pm} \textbf{0.02}$ & $0.33 \pm 0.01$ & $0.54 \pm 0.01$ & $-8.79 \pm 0.11$ & $-11.28 \pm 0.43$ \\
        & Fair-top-$k$ & $0.19 \pm 0.01$ & $1.05 \pm 0.02$ & $0.39 \pm 0.01$ & $\textbf{0.52} \bm{\pm} \textbf{0.01}$ & $-\textbf{5.05} \bm{\pm} \textbf{0.09}$ & $\textbf{0.47} \bm{\pm} \textbf{0.11}$ \\
        & Fair-top-$k$ \& DCL & $\textbf{0.15} \bm{\pm} \textbf{0.00}$ & $0.61 \pm 0.03$ & $0.41 \pm 0.01$ & $0.53 \pm 0.01$ & $5.36 \pm 0.10$ & $-\textbf{0.46} \bm{\pm} \textbf{0.46}$ \\
        \midrule
        \multirow{5}{*}{Credit} & Top-$k$ & $0.98 \pm 0.02$ & $1.24 \pm 0.04$ & $0.34 \pm 0.01$ & $0.46 \pm 0.01$ & $-8.98 \pm 0.19$ & $-20.18 \pm 0.49$ \\
        & Top-$k$ \& CL & $0.58 \pm 0.01$ & $0.90 \pm 0.05$ & $\textbf{0.30} \bm{\pm} \textbf{0.01}$ & $0.61 \pm 0.02$ & $-10.82 \pm 0.13$ & $-27.41 \pm 8.51$ \\
        & Top-$k$ \& DCL & $0.56 \pm 0.01$ & $1.02 \pm 0.05$ & $\textbf{0.30} \bm{\pm} \textbf{0.01}$ & $\textbf{0.43} \bm{\pm} \textbf{0.01}$ & $-10.90 \pm 0.15$ & $-19.16 \pm 0.96$ \\
        & Fair-top-$k$ & $0.48 \pm 0.01$ & {$1.33 \pm 0.04$} & $0.43 \pm 0.01$ & $0.45 \pm 0.01$ & $-\textbf{5.38} \bm{\pm} \textbf{0.07}$ & $-12.74 \pm 0.25$ \\
        & Fair-top-$k$ \& DCL & $\textbf{0.42} \bm{\pm} \textbf{0.01}$ & $\textbf{0.26} \bm{\pm} \textbf{0.02}$ & $0.43 \pm 0.01$ & $0.47 \pm 0.01$ & $-\textbf{5.51} \bm{\pm} \textbf{0.08}$ & $-\textbf{2.05} \bm{\pm} \textbf{0.24}$ \\
        \midrule
        \multirow{5}{*}{UCIcredit} & Top-$k$ & $0.75 \pm 0.02$ & $1.22 \pm 0.03$ & $\textbf{0.31} \bm{\pm} \textbf{0.01}$ & $0.49 \pm 0.01$ & $-7.21 \pm 0.18$ & $-24.01 \pm 0.62$ \\
         & Top-$k$ \& CL & $0.41 \pm 0.01$ & $0.53 \pm 0.04$ & $0.39 \pm 0.01$ & $0.65 \pm 0.01$ & $-7.08 \pm 0.10$ & $-2.91 \pm 0.43$ \\
        & Top-$k$ \& DCL & $0.42 \pm 0.02$ & $0.54 \pm 0.03$ & $0.38 \pm 0.01$ & $0.61 \pm 0.01$ & $-7.16 \pm 0.10$ & $-7.10 \pm 0.49$ \\
        & Fair-top-$k$ & $0.23 \pm 0.01$ & {$1.43 \pm 0.04$} & $0.45 \pm 0.01$ & $\textbf{0.46} \bm{\pm} \textbf{0.01}$ & $-\textbf{5.15} \bm{\pm} \textbf{0.08}$ & {$-5.13 \pm 0.23$} \\
        & Fair-top-$k$ \& DCL & $\textbf{0.18} \bm{\pm} \textbf{0.00}$ & $\textbf{0.40} \bm{\pm} \textbf{0.03}$ & $0.45 \pm 0.01$ & $0.58 \pm 0.01$ & $-\textbf{5.22} \bm{\pm} \textbf{0.08}$ & $\textbf{1.10} \bm{\pm} \textbf{0.19}$ \\
        \bottomrule
    \end{tabular}
    \caption{Average Recourse Cost, Fail to Recourse, and Higher Standard across three datasets on Logistic and MLP Models }
    \label{table2_model_performance}
\end{table*}

This subsection compares our proposed methods (Fair-top-$k$ and Dynamic continual learning) with the classical method (Top-$k$ with typical update) and the typical continual learning(Top-$k$ with continual learning update). The comparison shows similar trends across all three datasets, thus we report the results on the \textit{Credit} dataset and put the rest of them in Appendix.

Figure~\ref{fig:comparison: ours and baselines} illustrates the comparison of five different strategies: Fair-top-k labeling with a typical update (orange), Fair-top-$k$ labeling with a DCL update (red), Top-$k$ labeling with a DCL update (green), Top-$k$ labeling with a typical update (purple), and Top-$k$ labeling with continual learning (blue).

\textbf{Higher standard is eased among all settings}. The Higher Standard metric across the five methods shows that both Fair-top-$k$ solutions (orange and red lines) maintain values that are approximately 50\% better than other methods, indicating a more stable and consistent standard throughout the rounds. In contrast, the methods without the fair-top-$k$ policy (green, blue, and purple lines) exhibit a drastic decrease in both models. Furthermore, the fair-top-$k$ solutions also achieve a higher Test Acceptance Rate, as implied by the stable Higher Standard. Overall, Fair-top-$k$ labeling policy can ease the higher standard problem. 

\textbf{Recourse cost is significantly reduced and model accuracy is enhanced.}
In Table~\ref{table2_model_performance}, Fair-top-$k$ \& DCL consistently show lower Average Recourse Cost than other methods. For example, the cost of our methods is generally 70\% or lower than top-$k$.
This metric exhibits a strong correlation with the increase in the standard; higher standards correspond to higher recourse costs. However, the cost-invalidity trade-off \cite{RoCourseNet} persists, as observed in both Fail To Recourse and Average Recourse Cost. This means that a higher cost typically corresponds to a lower Fail To Recourse rate. Even within this trade-off, Fair-top-k \& DCL still achieve a significant cost reduction, while at most incurring a 14\% higher possibility of failing to recourse. On both models, Fair-top-$k$ policies (with and without DCL) achieve a strong, stable and sustained short-term balanced accuracy of approximately 90\% in the long term. For the MLP model, the top-$k$ with typical update method achieves the best short-term balanced accuracy in later rounds on the MLP model. This is attributed to its model quickly shifting to a higher standard early on, as seen in the test acceptance rate dropping to about 0 before 20 rounds, which then stabilized with minimal changes in later rounds.

\textbf{Continual learning reduces the recourse cost.}
As shown in Table~\ref{table2_model_performance}, methods employing a continual learning approach always achieve a lower cost compared to those using the same labeling policy without it. This advantage stems from continual learning's ability to retain knowledge from past models, ensuring the recourse actions remain stable and relevant over time.

\textbf{Our proposed methods provides tradeoff solutions in MLP models.}

The aforementioned three observations are also confirmed across other datasets, different recourse quality and ratio of recourse users. The summary of other factors is shown in Table~\ref{tab:fair-top-k} and the details are included in the appendix.

\setlength{\tabcolsep}{1mm}

\begin{table}[tb]
\centering
\fontsize{9pt}{11pt}\selectfont
\begin{tabular}{|c|c|c|c|c|}
\hline
\diagbox[width=2.6cm, height=1.0cm, font=\normalsize\bfseries]{Setting}{Situation}
& \makecell{Non-robust\\Model} 
& \makecell{Recourse\\Cost} 
& \makecell{Recourse\\Fail Rate}
& \makecell{Higher\\Standard}  \\
\hline
\makecell{Higher recourse\\quality} & $\uparrow$ & $\downarrow$ & $\downarrow$ & $\uparrow$ \\
\hline
\makecell{More\\recourse users} & $\uparrow$ & $\uparrow$ & $\uparrow$ & $\uparrow$ \\
\hline
\makecell{Larger k values} & $\downarrow$ & $\downarrow$ & $\downarrow$ & $\downarrow$ \\
\hline
\makecell{Inverse gamma\\cost weights} & -- & $\downarrow$ & -- & $\downarrow$ \\
\hline
\makecell{Higher ratio of\\original distri.} & $\uparrow$ & $\downarrow$ & -- & $\uparrow$ \\
\hline
\makecell{From logistics\\to MLP} & $\uparrow$ & $\uparrow$ & $\uparrow$ & $\downarrow$ \\
\hline
\end{tabular}
\caption{Effect of Different Settings on Fair-top-k Method.}
\label{tab:fair-top-k}
\end{table}

\section{Discussion, Conclusion, and Future Work}
\label{sec:discussion}

In this work, we study model evolution in the presence of recourse users under limited system resources. Our analysis and experiments indicate that recourse behavior naturally pushes the model toward a higher decision standard, thereby increasing the difficulty of future recourse actions—reflected in higher recourse costs and failure rates. To address these challenges, we proposed the Fair-top-$k$ strategy and a dynamic continual learning algorithm. Both methods significantly mitigate the problems identified and enhance model robustness, as evidenced by improved balanced accuracy over time in logistic models.

Nevertheless, this work represents only an initial step toward understanding the broader implications of recourse behavior in learning systems. There remain many open directions for further exploration, particularly in connection to fields such as strategic decision-making and social sciences. Below, we outline a few promising directions:

\paragraph{The prediction of the momentum in model evolution is still quite open.}
We showed that under certain conditions, model evolution tends to favor higher standards. However, it remains an open question whether one can theoretically characterize how recourse actions systematically influence the direction of model evolution. Specifically, what are the theoretical links between cost functions, recourse algorithms, and real-world constraints? On the other hand, the momentum in other configurations are still open, such as other models (e.g., random forests, complicated deep learning models, $\dots$ etc.), gaming framework, online \& real-time system, extreme resource constraints (in our experiments, when $k<.3$ all methods trigger higher standard and make short-term accuracy low).

\paragraph{The social impact on top-$k$ policy.}
The Top-$k$ strategy ultimately depends on a single score to determine classification results\footnote{In practice, even with complex deep learning models that consider multiple objectives, these objectives are often combined into one mixed objective function (e.g., via Lagrangian relaxation) to ensure differentiability.}. Consequently, users are driven to optimize a single “golden standard,” pushing the model toward stricter decision boundaries and reducing overall diversity. In our experiments, this reliance on a single metric not only causes model shifts towards a higher standard but can also generate social challenges such as user stress and anxiety~\cite{halko2017competitive,feri2013there}. Recent research from Purdue University indicates that an increasing number of content creators experience burnout or stop creating content on platforms like YouTube due to the competitive and stressful environment~\cite{thorne2023emotional}.

On the other hand, the cost of features is crucial in algorithmic recourse, yet it is rarely considered in model fitting and evolution. Although strategic learning~\cite{hardt2016strategic,levanon2021strategic} addresses this issue, its primary focus is often single-step accuracy rather than long-term dynamics. In reality, the cost function determines the direction of the data distribution’s shift, so inferring feature costs can help predict the trajectory of model evolution and even steer it intentionally. Additionally, the semantic meaning of features may play a significant role. For instance, should a video’s predicted quality rely more on its content-related characteristics or on socially driven metrics such as the number of likes? After all, careful system design and monitoring are probably essential to mitigate unintended consequences and ensure long-term stability and fairness~\cite{bell2024fairness}. Finally, we note that even if the cost function is unmeasurable or acts as a black box, one can still design strategies—such as the Fair-Top-$k$ approach—that accept diverse points and mitigate the pitfalls of focusing on a single score.

\section*{Acknowledgments}
Jie Gao would like to acknowledge funding support from NSF through CCF-2118953, DMS-2311064, DMS-2220271, IIS-2229876, and CNS-2515159. Hao-Tsung Yang would like to acknowledge funding support from NSTC through NSTC-114-2221-E-008-049 and NSTC-113-2221-E-008-086.

\bibliography{main_AIES}

\end{document}